\documentclass[runningheads]{llncs}
\usepackage{multirow}
\usepackage{amssymb}
\usepackage{algorithmic}
\usepackage{setspace}
\usepackage{graphicx}
\usepackage{lipsum}
\usepackage[ruled,lined]{algorithm2e}
\usepackage[dvipsnames]{xcolor}
\usepackage[hidelinks,colorlinks=true,linkcolor=blue,citecolor=blue]{hyperref}
\usepackage{printlen}

\uselengthunit{mm}
\begin{document}

\definecolor{commentcolor}{RGB}{110,154,155}   
\newcommand{\PyComment}[1]{\ttfamily\textcolor{commentcolor}{\# #1}}  
\newcommand{\PyCode}[1]{\ttfamily\textcolor{black}{#1}} 

\title{Group Distributionally Robust Knowledge Distillation}
%
%

\author{
Konstantinos Vilouras\inst{1} \and
Xiao Liu\inst{1,2} \and Pedro Sanchez\inst{1} \and
Alison Q. O'Neil\inst{1,2}  \and
Sotirios A. Tsaftaris\inst{1}
}
\authorrunning{K. Vilouras et al.}
\institute{
School of Engineering, University of Edinburgh, Edinburgh EH9 3FB, United Kingdom\\
\email{\{konstantinos.vilouras,xiao.liu,pedro.sanchez,s.tsaftaris\}@ed.ac.uk} \and
Canon Medical Research Europe Ltd., Edinburgh EH6 5NP, United Kingdom\\
\email{alison.oneil@mre.medical.canon}
}

%
%
\maketitle              
\setcounter{footnote}{0}
\begin{abstract}
Knowledge distillation enables fast and effective transfer of features learned from a bigger model to a smaller one. However, distillation objectives are susceptible to \emph{sub-population shifts}, a common scenario in medical imaging analysis which refers to groups/domains of data that are underrepresented in the training set. For instance, training models on health data acquired from multiple scanners or hospitals can yield subpar performance for minority groups. In this paper, inspired by distributionally robust optimization (DRO) techniques, we address this shortcoming by proposing a group-aware distillation loss. During optimization, a set of weights is updated based on the per-group losses at a given iteration. This way, our method can dynamically focus on groups that have low performance during training. We empirically validate our method, \textit{GroupDistil} on two benchmark datasets (natural images and cardiac MRIs) and show consistent improvement in terms of worst-group accuracy.
\end{abstract}
\section{Introduction}

The rapid success of deep learning can be largely attributed to the availability of both large-scale training datasets and high-capacity networks able to learn arbitrarily complex features to solve the task at hand. Recent practices, however, pose additional challenges in terms of real-world deployment due to increased complexity and computational demands. Therefore, developing lightweight versions of deep models without compromising performance remains an active area of research.

\noindent \textbf{Knowledge distillation} To this end, knowledge distillation is a promising technique aiming to guide the learning process of a small model (\textit{student}) using a larger pre-trained model (\textit{teacher}). This guidance can be exhibited in various forms; for example, Hinton et al.\ \cite{hinton2015distilling} propose to match the soft class probability distributions predicted by the teacher and the student. Tian et al.\ \cite{tian2019contrastive} augment this framework with a contrastive objective to retain the dependencies between different output dimensions (since the loss introduced in \cite{hinton2015distilling} operates on each dimension independently). On the contrary, Romero et al.\ \cite{romero2014fitnets} consider the case of matching intermediate teacher and student features (after projection to a shared space since the original dimensions might be mismatched). Another interesting application involves distilling an ensemble of teachers into a single student model, which has been previously explored in the case of histopathology \cite{tellez2018whole} and retinal \cite{ju2021relational} images.

\noindent \textbf{Distillation may lead to poor performance} Although knowledge distillation has been widely used as a standard approach to model compression, our understanding of its underlying mechanisms remains underexplored. Recently, Ojha et al.\ \cite{ojha2022knowledge} showed, through a series of experiments, that a distilled student exhibits both useful properties such as invariance to data transformations, as well as biases, e.g.,\ poor performance on rare subgroups inherited from the teacher model. Here, we attempt to tackle the latter issue which can be especially problematic, for instance, for medical tasks where we have access to only a few data for certain groups in the training set. Note that there also exist concurrent works \cite{wang2022robust} similar to ours that consider the case of distillation on long-tailed\footnote{i.e., on sets where the majority of data comes from only a few classes} datasets.

In this paper, we consider the task of knowledge distillation under a common type of distribution shift called \emph{sub-population shift} where both training and test sets overlap, yet per-group proportions differ between sets. In sum, our contributions are the following:
\begin{itemize}
    \item We propose a simple, yet effective, method that incorporates both the original distillation objective and also group-specific weights that allow the student to achieve high accuracy even on minority groups.
    \item We evaluate our method on two publicly available datasets, i.e., on natural images (Waterbirds) and cardiac MRI data (M\&Ms) and show improvements in terms of worst-group accuracy over the commonly used knowledge distillation loss of \cite{hinton2015distilling}.
\end{itemize}

\section{Methodology}
\noindent \textbf{Preliminaries} Let $D$ denote the data distribution from which triplets of data instances $x \in \mathcal{X}$, labels $y \in \mathcal{Y}$ and domains $d \in \mathcal{D}$ are sampled, respectively. Let also the teacher $f_T: \mathcal{X} \rightarrow \mathcal{Z}_T$ and student $f_S: \mathcal{X} \rightarrow \mathcal{Z}_S$ map input samples to class logits. The vanilla knowledge distillation loss introduced in Hinton et al.\ \cite{hinton2015distilling} is defined as
\begin{equation}
\mathcal{L}_{KD} = (1-\alpha) \cdot H(y, \sigma (z_S)) + \alpha \tau^2\cdot D_{KL}(\sigma (z_T / \tau), \sigma (z_S / \tau)),
\label{eq:eq1}
\end{equation}
where $H(p,q)=-\mathbb{E}_p[\log q]$ refers to cross-entropy, $D_{KL}(p,q)=H(p,q)-H(p)$ is the Kullback-Leibler divergence metric, $\sigma$ is the softmax function, $\tau$ is the temperature hyperparameter and $\alpha$ the weight that balances the loss terms.

\begin{figure}[t!]
    \centering
    \includegraphics[width=1\textwidth,, trim = 1cm 18cm 0cm 2cm, clip]{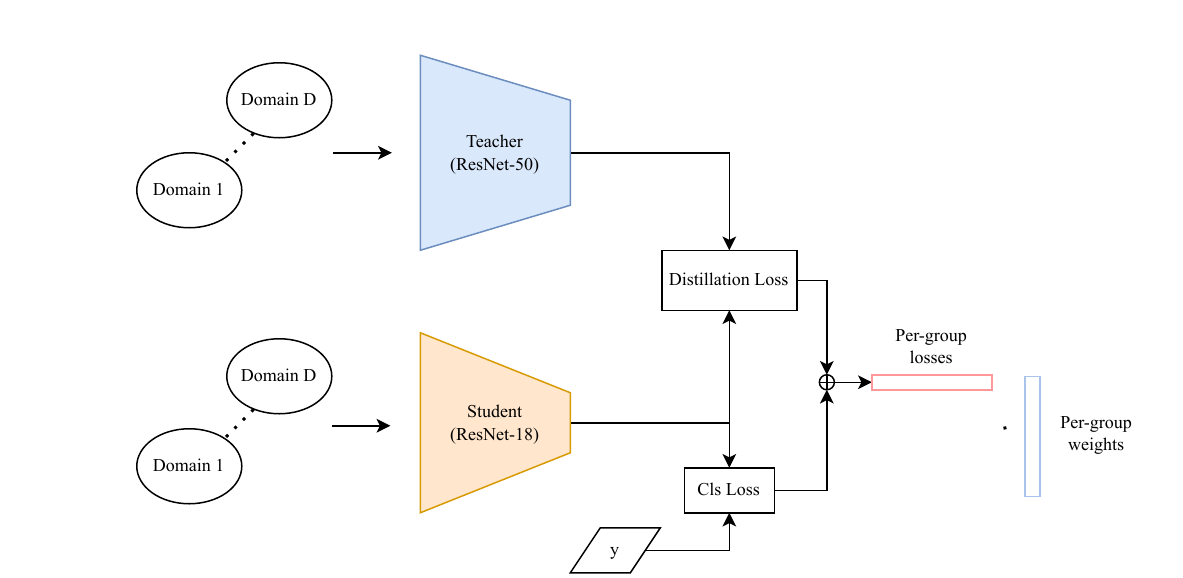}
    \caption{Overview of the proposed method \textit{GroupDistil}. Data from multiple domains $d=1,...,|\mathcal{D}|$ are fed to both the teacher and student to produce class logits. A soft version of the probability distribution over classes induced by both models is used to calculate the distillation loss, whereas the hard version ($\tau=1$) of the student's probability distribution is used for the classification loss. The final loss is calculated as the dot product between the per-group losses and the group weights. This way, more emphasis is placed on domains where the student has low performance.} 
    \label{fig2}
\end{figure}

In our case, inspired by the groupDRO algorithm of Sagawa et al.\ \cite{sagawa2019distributionally}, we propose an alternative objective that incorporates group-specific weights (termed \textit{GroupDistil}). An overview of our method is presented in Figure \ref{fig2}. The final form of the proposed loss function is shown in Equation \ref{eq:eq2}, where $w_d$ and $\mathcal{L}_{KD}^d$ refer to the weight and the dedicated distillation loss for group $d$, respectively. This allows the student to emphasize domains where performance remains low (e.g., rare subgroups within the training set) by upweighting their contribution to the overall loss. In that sense, the final optimization objective can be divided into two steps, i.e., first performing exponentiated gradient ascent on group weights and then minibatch stochastic gradient descent on student's weights. Algorithm \ref{algo:your-algo} provides a full description of the proposed method. The main difference with the groupDRO algorithm \cite{sagawa2019distributionally} is the choice of loss function, i.e., we use the distillation loss instead of cross-entropy, and it is highlighted in Algorithm \ref{algo:your-algo} in blue.
\begin{equation}
\mathcal{L}_{GroupDistil} = \sum_{d=1}^\mathcal{|D|} w_d\cdot \mathcal{L}_{KD}^d,
\label{eq:eq2}
\end{equation}

\begin{algorithm}[!t]
    \DontPrintSemicolon
    \SetKwInput{Input}{Input}
    \SetKwInOut{Output}{Output}
    \SetKwComment{Comment}{// }{ }

    \Input{Data distribution for a given domain $P_d$; Learning rate $\eta_\theta$; Group weight step size $\eta_w$}
    Initialize (uniform) group weights $w^{(0)}$ and student weights $\theta_s^{(0)}$\\
    \For{$t=1,...,T$}{
        $d \sim \mathcal{U}(1, ..., |\mathcal{D}|)$ \Comment*[r]{Draw a random domain}
        $(x, y) \sim P_d$ \Comment*[r]{Sample (data, labels) from domain $d$}
        \color{blue}
        $\ell \leftarrow \mathcal{L}_{KD}^d(x,y)$ \Comment*[r]{Calculate distillation loss (Eqn \ref{eq:eq1})}
        \color{black}
        $w' \leftarrow w^{(t-1)}$; $w'_d\leftarrow w'_d\cdot \exp(\eta_w \ell)$ \Comment*[r]{Update weight for domain $d$}
        $w^{(t)} \leftarrow w' / \sum_{d'}w'_{d'}$ \Comment*[r]{Normalize new weights}
        $\theta_s^{(t)}\leftarrow \theta_s^{(t-1)} - \eta_\theta w_d^{(t)}\nabla \ell$ \Comment*[r]{Optimize student weights via SGD}
        
    }
\caption{\emph{GroupDistil} method}
\label{algo:your-algo}
\end{algorithm}

\section{Datasets}
\subsection{Waterbirds}
\begin{figure}[h!]
    \centering
    \includegraphics[width=1\textwidth]{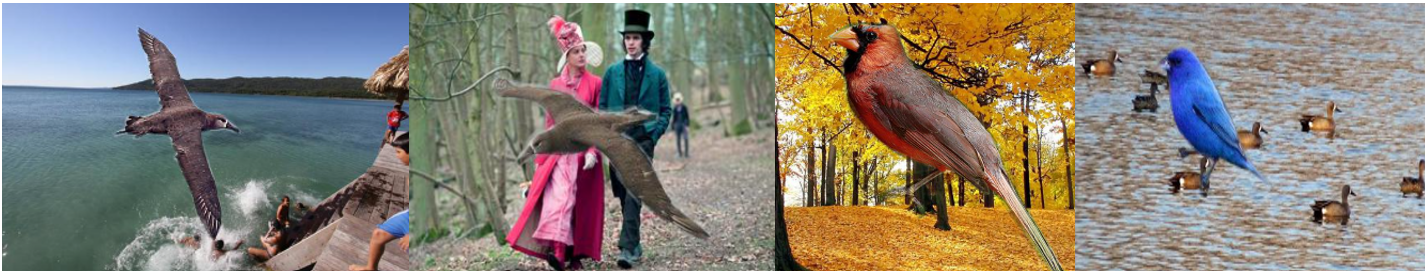}
    \caption{Examples of images from Waterbirds dataset. Left: waterbird on water. Middle left: waterbird on land. Middle right: landbird on land. Right: landbird on water.} 
    \label{waterbirds}
\end{figure}

Waterbirds \cite{koh2021wilds,sagawa2019distributionally} is a popular benchmark used to study the effect of sub-population shifts. More specifically, it contains instances from all possible combinations of $\mathcal{Y} =$ \{landbird, waterbird\} labels and $\mathcal{D} =$ \{land, water\} backgrounds (4 domains in total), respectively. However, it is collected in a way such that uncommon pairs (i.e.,\ landbirds on water and waterbirds on land) occur less frequently, thus creating an imbalance. Also, note that the level of imbalance in the training set is different than that of the test set. Training, validation and test sets consist of 4795, 1199 and 5794 samples, respectively. In our setup, we first resize images to fixed spatial dimensions ($256\times 256$), then extract the $224\times 224$ center crop and normalize using Imagenet's mean and standard deviation. A few representative samples of this dataset are depicted in Figure \ref{waterbirds}.

\subsection{M\&Ms}
\begin{figure}[h!]
    \centering
    \includegraphics[width=1\textwidth]{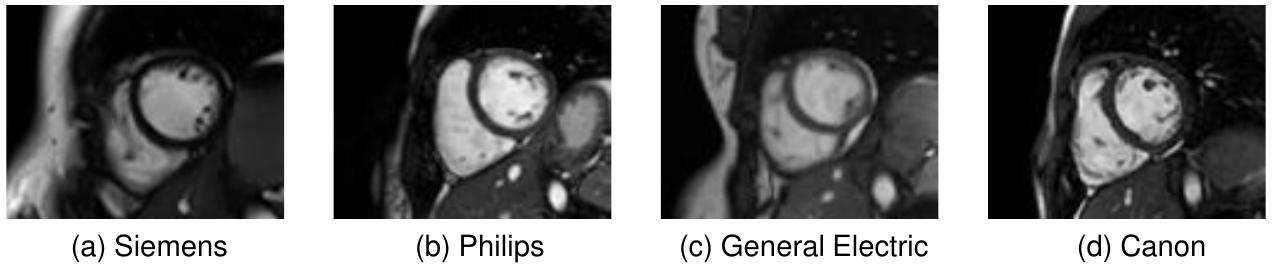}
    \caption{Examples of cardiac MRI images from M$\&$Ms dataset \cite{mnms}.} 
    \label{mnmsdata}
\end{figure}

The multi-centre, multi-vendor and multi-disease cardiac image segmentation (M$\&$Ms) dataset \cite{mnms} contains 320 subjects. Subjects were scanned at 6 clinical centres in 3 different countries using 4 different magnetic resonance scanner vendors (Siemens, Philips, GE, and Canon) i.e., domains A, B, C and D. For each subject, only the end-systole and end-diastole phases are annotated. Voxel resolutions range from $0.85\times 0.85\times 10$ mm to $1.45\times 1.45\times 9.9$ mm. Domain A contains 95 subjects. Domain B contains 125 subjects. Both domains C and D contain 50 subjects. We show example images of the M$\&$Ms data in Fig. \ref{mnmsdata}. Note that, while the dataset was originally collected for the task of segmentation, we instead use it here for classification.

\section{Experimental Results}
We now discuss our results.
 First, we briefly discuss hyperparameter choices made on a per-dataset basis. In each case, we compare the performance of three types of \emph{students}: an individual student that was trained with a group robustness method \cite{sagawa2019distributionally} from scratch (\emph{groupDRO} hereafter), a distilled student trained with the objective of Equation \ref{eq:eq1} (\emph{KD} hereafter) and also a distilled student trained with our proposed \emph{GroupDistil} method as described in Equation \ref{eq:eq2} and Algorithm \ref{algo:your-algo} (\emph{GroupDistil} hereafter).

\noindent \textbf{Waterbirds} We consider the case of distilling the knowledge from a ResNet-50 teacher model pre-trained with groupDRO to a ResNet-18 student (shorthanded as R$50\rightarrow$ R$18$). Note that our proposed method does not impose any restrictions on either the choice of model architectures or the teacher's pre-training strategy (groupDRO was merely chosen to ensure that the teacher learns robust features on the given training set). For our objective, we use the following hyperparameters: batch size $=128$, total number of epochs $=30$, Adam optimizer with learning rate $\eta_\theta=10^{-4}$, temperature $T=4$, $\alpha=0.9$ and group weight step size $\eta_w = 0.01$. Final results on the official test set are depicted in Table \ref{tab1}.

\begin{table}[h!]
\caption{Distillation results on Waterbirds dataset. First two rows show performance of each model trained from scratch. The last two rows refer to vanilla knowledge distillation (KD) and our proposed method (GroupDistil), respectively. In those cases, we fix the teacher (1st row) and initialize each R18 model with ImageNet-pretrained weights. We report both the average (2nd column) and worst-group (3rd column) accuracy on the official test set. Final results have been averaged across 3 random seeds.}\label{tab1}
\centering
\begin{tabular}{|c|c|c|}
\hline
Setup & Adjusted avg acc. (\%) & Worst-group acc. (\%)\\
\hline
groupDRO (R50) & 90.9 & 86.8 \\
groupDRO (R18) & 86.0 & 79.6 \\
\hline
KD (R$50\rightarrow$ R$18$) & 95.2 $\pm$ 0.1 & 78.0 $\pm$ 1.5 \\
\textit{GroupDistil} (R$50\rightarrow$ R$18$) & 86.6 $\pm$ 0.8 & \textbf{81.7} $\pm$ 0.8 \\
\hline
\end{tabular}
\end{table}

Our \emph{GroupDistil} method shows consistent improvement ($>$ 2\%) compared to the knowledge distillation objective (KD) of \cite{hinton2015distilling} in terms of worst-group accuracy. This implies that our distilled student does not rely on spurious attributes (background information) to classify each type of bird. This is also evident from the fact that worst-group accuracy in our case does not deviate much (approx. 5\%) from the average accuracy in the test set. Note that our distilled student even outperforms a ResNet-18 model trained from scratch using groupDRO, showing that smaller models can largely benefit from larger ones in this setup.

\noindent \textbf{M\&Ms} In the case of M\&Ms, since there is no official benchmark for studying sub-population shifts, we divide the available data in training and test sets as follows: First, we consider a binary classification task using the two most common patient state classes\footnote{This choice eliminates Domain B due to lack of available data.}, i.e., hypertrophic cardiomyopathy (HCM) and healthy (NOR). Then, we conduct our experiments using two types of splits as presented in Table \ref{mnms_split}. Splits are defined in such a way that allows us to measure performance on a specific type of scanner (domain) that is underrepresented in the training set. Thus, for the first (resp.\ second) split, we keep only 2 patients per class from Domain C (resp.\ D) in the train set, and the rest in the test set. Note that, for fair comparison, we ensure that the test set is always balanced.

\begin{table}[h!]
\caption{M\&Ms dataset. For each type of experiment, we show the total number of patients per dataset (train or test), domain (A, C, or D) and class (HCM or NOR), respectively. Note that we extract 20 2D frames from each patient for both training and testing.}\label{mnms_split}
\centering
\begin{tabular}{|c|c|c|c|}
\hline
& & Train & Test \\
\cline{3-4}
Experiment & Class & Domains (A/C/D) & Domains (A/C/D)\\
\hline
\multirow{2}{*}{Split 1} & HCM & 25/2/10 & -/3/- \\
& NOR & 21/2/14 & -/3/- \\
\hline
\multirow{2}{*}{Split 2} & HCM & 25/5/2 & -/-/8 \\
& NOR & 21/11/2 & -/-/8 \\
\hline
\end{tabular}
\end{table}

For this dataset, we use a ResNet-18 (pre-trained with groupDRO) as teacher and a DenseNet-121 as a student model (R$18\rightarrow$ D$121$ setup). As a pre-processing step, we first extract a random 2D frame from each 4D input (in total, we extract 20 frames from each patient per epoch) and then apply data augmentations such as intensity cropping, random rotation and flip. The input for each model is a $224\times 224$ crop from each frame. We also used the following hyperparameters: batch size $=128$, total number of epochs $=5$, Adam optimizer with learning rate $\eta_\theta=5\cdot10^{-4}$, temperature $T=4$, $\alpha=0.9$ and group weight step size $\eta_w = 0.01$. Results for the first split are shown in Table \ref{tab3}, whereas for the second split in Table \ref{tab4}.

As in Waterbirds, similar observations can be made for the M\&Ms dataset. It is clear that in both types of splits, our method reaches the highest accuracy. Also note that the teacher and the student trained from scratch have low performance on the test set, indicating the challenging nature of this dataset; yet, distilled students can significantly outperform them. The results for KD method exhibit high variance (possibly due to the limited number of available data), indicating that it could be unstable in this setup. On the contrary, our method remains fairly robust and shows consistent performance improvements over the rest of the methods.

\begin{table}[t!]
\caption{Distillation results on M\&Ms dataset, 1st split (testing on Domain C). First two rows show performance of each model trained from scratch. The last two rows refer to vanilla knowledge distillation (KD) and our proposed method (GroupDistil), respectively. We report the average accuracy on the test set. Final results have been averaged across 5 random seeds.}\label{tab3}
\centering
\begin{tabular}{|c|c|}
\hline
Setup & Accuracy (\%)\\
\hline
groupDRO (R18) & 53.3\\
groupDRO (D121) & 46.7\\
\hline
KD (R$18\rightarrow$ D$121$) & 55.8 $\pm$ 7.0\\
\textit{GroupDistil} (R$18\rightarrow$ D$121$) & \textbf{62.8} $\pm$ 4.9\\
\hline
\end{tabular}
\end{table}

\begin{table}[t!]
\caption{Distillation results on M\&Ms dataset, 2nd split (testing on Domain D). First two rows show performance of each model trained from scratch. The last two rows refer to vanilla knowledge distillation (KD) and our proposed method (GroupDistil), respectively. We report the average accuracy on the test set. Final results have been averaged across 5 random seeds.}\label{tab4}
\centering
\begin{tabular}{|c|c|}
\hline
Setup & Accuracy (\%)\\
\hline
groupDRO (R18) & 62.2\\
groupDRO (D121) & 56.2\\
\hline
KD (R$18\rightarrow$ D$121$) & 64.2 $\pm$ 7.4\\
\textit{GroupDistil} (R$18\rightarrow$ D$121$) & \textbf{66.5} $\pm$ 2.8\\
\hline
\end{tabular}
\end{table}

\section{Conclusion and Future Work}
In this paper we consider the task of knowledge distillation under a challenging type of distribution shift, i.e., sub-population shift. We showed that adding group-specific weights to a popular distillation objective provides a significant boost in performance, which even outperforms the same student architecture trained from scratch with a group robustness method in terms of worst-group accuracy. We also made sure that our proposed method remains fairly general, allowing arbitrary combinations of teacher-student models.

A limitation of our work is the fact that we assume access to fully labeled data, i.e., with both label and domain annotations, which is restrictive in practice. Therefore, for future work, we plan to investigate methods that infer domains directly from data as in \cite{creager2021environment,zhang2022correct}.

\section{Acknowledgements}
This work was supported by the University of Edinburgh, the Royal Academy of Engineering and Canon Medical Research Europe by a PhD studentship to Konstantinos Vilouras. S.A.\ Tsaftaris also acknowledges the support of Canon Medical and the Royal Academy of Engineering and the Research Chairs and Senior Research Fellowships scheme (grant RCSRF1819\textbackslash 8\textbackslash 25), and the UK’s Engineering and Physical Sciences Research Council (EPSRC) support via grant EP/X017680/1.  

\bibliography{bibliography.bib}

\end{document}